\documentclass{article} 

\usepackage{Apriel_1_5_15B}
\usepackage{comment}
\usepackage{multicol}

\usepackage[utf8]{inputenc} 
\usepackage[T1]{fontenc}    
\usepackage{newunicodechar}
\usepackage{enumitem}
\newunicodechar{ }{\,} 
\usepackage[colorlinks=true,    
            linkcolor=blue,     
            citecolor=blue,      
            urlcolor=blue]   
           {hyperref}
\usepackage{url}            
\usepackage{booktabs}       
\usepackage{amsfonts}       
\usepackage{nicefrac}       
\usepackage{microtype}      
\usepackage{xcolor}         
\usepackage{makecell}
\usepackage{tikz}
\usepackage{pgfplots}
\usepackage{graphicx}
\usepackage{subcaption}
\usepackage{multirow}
\usepackage[table]{xcolor}
\usepackage{caption}

\newcommand{\modelname}{{Apriel-1.5-15B-Thinker}}

\pgfplotsset{compat=1.18}
\pgfplotsset{scaled ticks=false}

\usepackage[
  backend=biber,
  style=numeric,
  sorting=none
]{biblatex}

\bibliography{Apriel_Nemotron_15B}

\title{Apriel-1.5-15B-Thinker: Mid-training is all you need}

\begin{document}
\maketitle
\begin{figure}[h]
    \centering
    \includegraphics[width=0.2\textwidth]{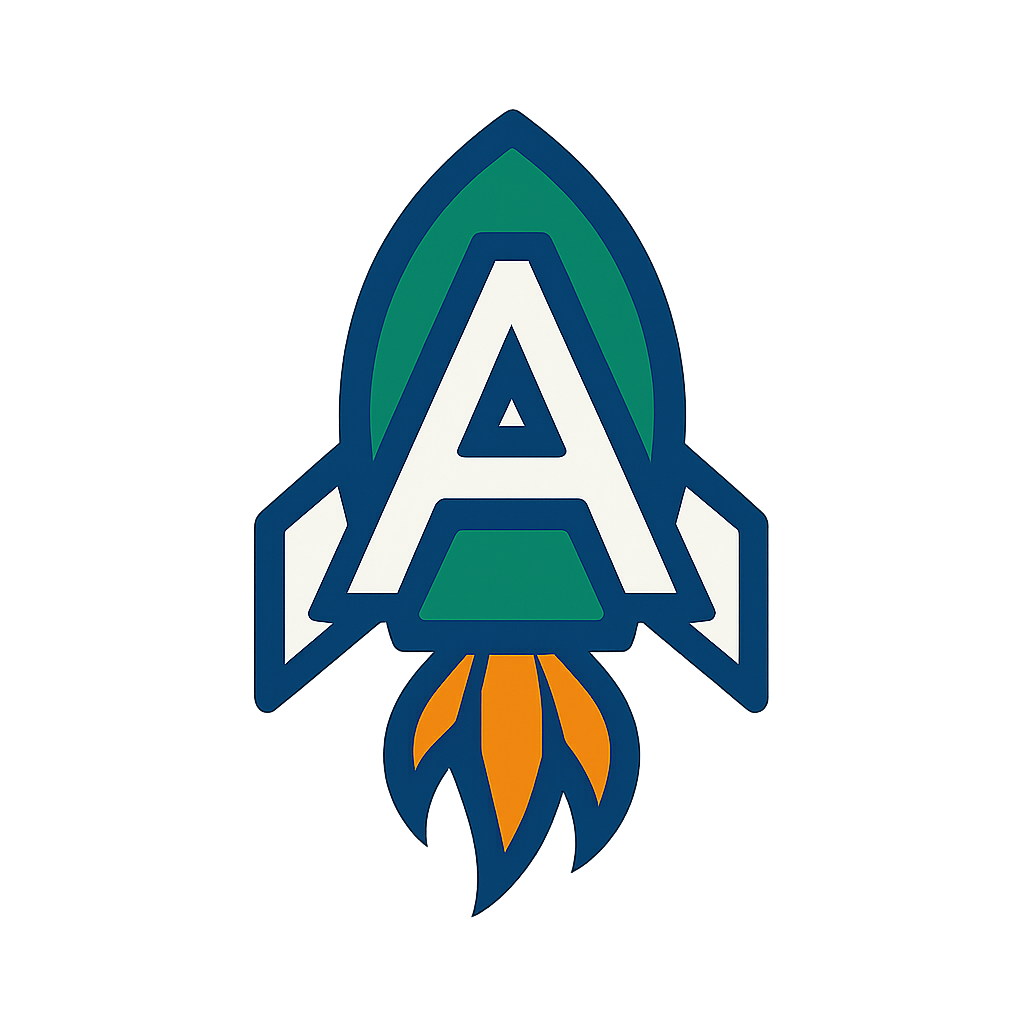}
\end{figure}

\begin{abstract}
  We present \modelname, a 15-billion parameter open-weights multimodal reasoning model that achieves frontier-level performance through training design rather than sheer scale. Starting from Pixtral-12B, we apply a progressive three-stage methodology: (1) depth upscaling to expand reasoning capacity without pretraining from scratch, (2) staged continual pre-training that first develops foundational text and vision understanding, then enhances visual reasoning through targeted synthetic data generation addressing spatial structure, compositional understanding, and fine-grained perception, and (3) high-quality text-only supervised fine-tuning on curated instruction-response pairs with explicit reasoning traces spanning mathematics, coding, science, and tool use. Notably, our model achieves competitive results without reinforcement learning or preference optimization, isolating the contribution of our data-centric continual pre-training approach. On the Artificial Analysis Intelligence Index, \modelname\ attains a score of 52, matching DeepSeek-R1-0528 despite requiring significantly fewer computational resources. Across ten image benchmarks, its performance is on average within five points of Gemini-2.5-Flash and Claude Sonnet-3.7, a key achievement for a model operating within single-GPU deployment constraints. Our results demonstrate that thoughtful mid-training \footnote{We define mid-training as a combination of the continual pretraining and SFT stages} design can close substantial capability gaps without massive scale, making frontier-level multimodal reasoning accessible to organizations with limited infrastructure. We release the model checkpoint, all training recipes, and evaluation protocols under the MIT license to to advance open-source research.

\end{abstract}

\begin{figure}[h]
    \centering
    \captionsetup{font=small}
    \includegraphics[width=1.0\textwidth]{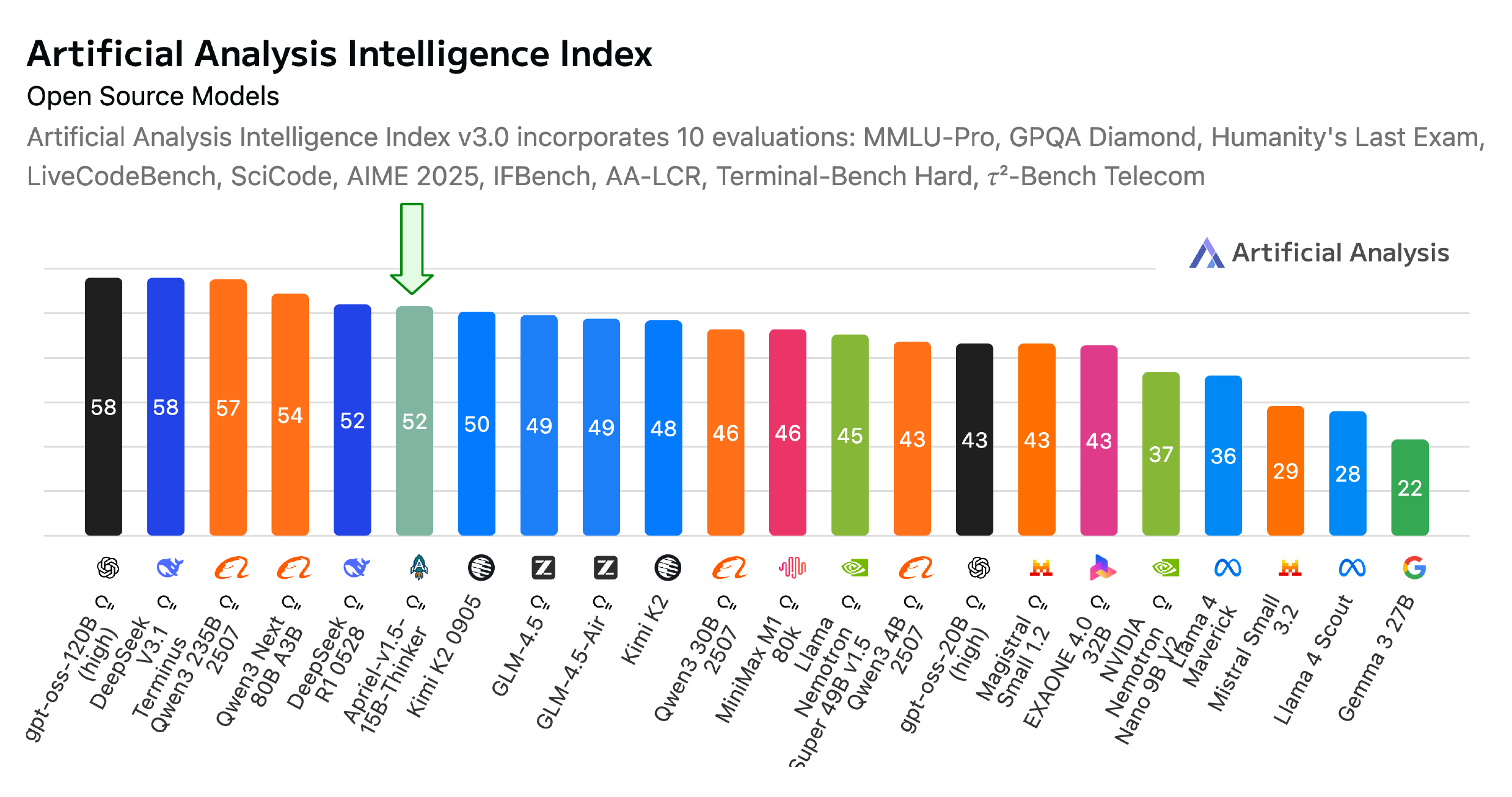}
    \caption{\modelname\ compared to the best open source LLMs on the Artificial Analysis Intelligence Index.}
    \label{fig:aa-open-source}
\end{figure}

\section{Introduction}

Large language models (LLMs) continue to advance rapidly across general capability, long-context reasoning, and multimodal understanding. Open-weight families such as Qwen~\cite{yang2024qwen2,yang2024qwen2_5} and Llama~\cite{grattafiori2024llama3herdmodels, llama4} have demonstrated strong, scalable baselines, while proprietary systems like Gemini~\cite{gemini15,gemini25} and Claude~\cite{claude35sonnet} have pushed frontier performance across complex reasoning and multimodal tasks. Recent “reasoning-first” training approaches, exemplified by DeepSeek-R1~\cite{deepseekai2025deepseekr1incentivizingreasoningcapability}, reveal that careful data curation and training strategy can unlock sophisticated chain-of-thought competence without relying solely on extreme scale. Yet despite these advances, a fundamental tension persists between capability and accessibility, as these insights have not fully addressed the challenges facing real-world applications.

Two critical barriers remain for widespread adoption. First, organizations requiring on-premises or air-gapped deployments for privacy and compliance need compact models with predictable resource footprints that can operate within strict infrastructure constraints. Second, the cost profile spanning both training and inference often becomes the decisive factor in whether frontier-level capability can be deployed at production scale. These practical considerations raise a fundamental question: \emph{Can a compact, open, multimodal model achieve frontier-level reasoning while remaining economical to train and deploy?}

This work introduces \textbf{\modelname}, a 15B-parameter open-weights multimodal reasoning model designed with that guiding question in mind. Our approach centers on the mid-training/continual pretraining phase, where both data selection and staged presentation exert a strong influence on downstream reasoning. Concretely, the training corpus spans curated pretraining-style corpora, diverse web-style text and images, reasoning-rich samples, and a mix of verified and unverified synthetic data, all introduced through a staged curriculum.  Our core innovation lies in a progressive, cost-effective multimodal training pipeline that effectively scales reasoning capabilities across text and vision through three carefully orchestrated stages:

\textbf{(1) Integrated Multimodal Architecture:}
Beginning with Pixtral-12B~\cite{agrawal2024pixtral12b} as our foundation, we expand it to a model size capable of advanced reasoning across modalities, without requiring pretraining from scratch. 

\textbf{(2) Staged Multimodal Continual Pretraining (CPT):} We adopt a two-phase CPT strategy. The first phase develops foundational text reasoning and broad multimodal capabilities, while the second enhances visual reasoning through synthetic data targeting spatial structure, compositional understanding, and fine-grained perception. This staged progression enables balanced strengthening of both modalities and provides a stable foundation for subsequent training stages, even when later stages emphasize a narrower set of modalities.

\textbf{(3) High-Quality Supervised Fine-Tuning (SFT):} We curate a diverse, high-quality, and high-signal set of samples for supervised fine-tuning. Each response includes explicit reasoning traces, enabling the model to learn transparent thought processes. Coupled with the strong base model, this yields frontier-level performance across a broad range of reasoning benchmarks without requiring additional post-training. 

Given computational constraints, the current release focuses on maximizing the potential of the base model through mid-training, without employing reinforcement learning or preference optimization. This design choice also allows for a clearer assessment of the contribution of the mid-training recipe itself to the overall performance of the model.

The result is a compact model tailored to enterprise-friendly deployment constraints (memory, latency, throughput) while still achieving frontier-level reasoning. \modelname\ attains a score of 52 on the Artificial Analysis Intelligence Index, matching DeepSeek-R1-0528~\cite{deepseekr1} despite requiring significantly fewer computational resources. Across ten multimodal benchmarks, the model demonstrates competitive performance, averaging only 5 points behind Gemini-2.5-Flash and Claude Sonnet-3.7~\cite{gemini25,claude35sonnet}, a remarkable achievement for a 15B parameter model operating within single-GPU deployment constraints. These empirical results provide compelling evidence that thoughtful continual pre-training with heterogeneous synthetic signals, applied to a compact architecture, can close substantial capability gaps without massive scale or expensive RL pipelines. By releasing this open, compact, multimodal reasoning model that approaches the frontier, we aim to catalyze research on mid-training curricula and lower operational barriers for privacy-preserving, cost-aware deployments across diverse organizational contexts.

\textbf{Summary of Contributions.} This work advances the state of efficient multimodal reasoning through four interconnected contributions that challenge conventional assumptions about scale, cost, and accessibility:

\begin{itemize}[noitemsep, topsep=0pt, leftmargin=*]
 \item \textbf{Compact and Deployable Frontier-level Models:} We show that relatively small models can achieve frontier-level performance, narrowing the gap to leading proprietary systems through training design rather than sheer parameter count. Their modest compute and memory footprint further makes them practical for on-premises deployment in constrained environments. 

 \item \textbf{Efficient and Democratized Scaling:} Our staged continual pretraining recipe strengthens both textual and visual reasoning while remaining feasible under realistic budgets. Techniques such as depth upscaling (capacity expansion without pretraining from scratch), selective loss computation, and checkpoint averaging improve efficiency and stability. This maximizes the potential of the base model and demonstrates that approaching state-of-the-art performance is not restricted to organizations with tens of thousands of GPUs.


 \item \textbf{Comprehensive Cross-Modal Evaluation:} We demonstrate strong results across text reasoning benchmarks (e.g. AIME’25: 88\%, IFBench: 62\%, $\tau^2$-Bench Telecom: 68\%) and multimodal tasks (e.g. MMMU: 70.2\%, MathVista: 75.5\%, CharXiv: 88.2\%). These results underscore broad reasoning competence across domains, supported by both internal evaluation and third-party validation.

 \item \textbf{Open-Source Compact Multimodal Foundation Model:} To our knowledge, this is the first openly released compact multimodal reasoning model operating at frontier level. We provide weights, training recipes, and evaluation artifacts under a permissive license, democratizing access and enabling reproducibility and further study

\end{itemize}

We structure the report as follows: Section~\ref{sec:arch} introduces the multimodal architecture integration, including depth upscaling and cross-modal training procedures. Section~\ref{sec:cpt} presents our staged continual pretraining, detailing the data and training methodology across the foundational reasoning, image understanding phase and the visual reasoning phase. Section~\ref{sec:sft} describes our high-quality data curation and training for supervised fine-tuning. Section~\ref{sec:eval} describes our evaluation methodology for text and image benchmarks, incorporating internal validation and third-party assessment. Section~\ref{Benchmarking} reports comprehensive evaluations across text and multimodal benchmarks, along with additional analysis. Finally, Section~\ref{sec:conclusion} concludes the report with possible directions for future work. 

\section{Architecture and Model Upscaling} \label{sec:arch}

\paragraph{Base Model}
To enable multimodal capabilities in a compute efficient manner, we build on Pixtral-12B-Base-2409 \cite{agrawal2024pixtral12b}\footnote{https://huggingface.co/mistralai/Pixtral-12B-Base-2409
. We used a version from Unsloth, which is no longer available at https://huggingface.co/unsloth
 as of this writing.}. Pixtral follows the LLaVA architecture \cite{liu2023llava}, consisting of a vision encoder connected to a multimodal decoder through a two-layer fully connected projection network.

\paragraph{Depth Upscaling}
Following the approach adopted in Apriel-Nemotron-15B-Thinker \cite{radhakrishna2025aprielnemotron15bthinker}, we first upscale the base model via depth upscaling to balance compute, latency, and performance, while maintaining deployability on a single high-end GPU. To upscale the multimodal model, we first upscale the decoder by increasing the number of hidden layers from 40 to 48, training on a large corpus of text tokens. Half of these tokens serve as replay data, and the rest are drawn from diverse domains including high-quality web content, technical literature, mathematical problem sets, programming code, and StackExchange discussions. 

\paragraph{Projection network realignment}
Next, the projection network is realigned by training on data from image captioning datasets, multimodal instruction-response pairs, and document understanding scenarios. During this stage, the vision encoder and the decoder remain frozen. 

\paragraph{Training Setup}
Both depth upscaling and projection network realignment were trained with a sequence length of 8192 (with sequence packing) and a learning rate of 5e-5 with linear decay. The weights of six equispaced intermediate checkpoints from the depth upscaling stage were averaged in equal proportions before projection network realignment. The final checkpoint obtained from the projection network realignment stage was used for subsequent stages of training.


\section{Continual Pretraining (CPT)} \label{sec:cpt}
To strengthen the foundational capabilities of the base model, after scaling up the model we further enhance its textual and visual reasoning capabilities with multimodal continual pretraining (CPT). The CPT process is divided into two stages: the first focuses on enhancing the model’s textual reasoning and image understanding capabilities, while the second aims at further improving its visual reasoning capabilities. The two stages are described in detail below.

\subsection{CPT Stage 1}

\paragraph{Foundational Reasoning and Multimodal Data}
The first stage involves training on a dataset that comprises of $50\%$ text-only tokens covering mathematical and scientific reasoning, coding tasks, and general knowledge; $20\%$ tokens replayed from the decoder upscaling stage; and $30\%$ multimodal tokens drawn from data on document understanding, chart understanding and reasoning, image captioning, long-form image descriptions, OCR-related tasks, and reasoning over mathematical and logical problems in visual contexts. 

\paragraph{Training Setup}
Since this stage involved data addressing foundational vision capabilities of the model, the vision encoder, projection network, and decoder were kept unfrozen. The training was performed at a sequence length of 32768 (with sequence packing) and a learning rate of 5e-5 with cosine decay and 10\% warmup. Loss was computed on all the tokens in the sequence. The weights of three equispaced intermediate checkpoints were averaged in equal proportions to form the final checkpoint from this stage. 

\subsection{CPT Stage 2}

\paragraph{Targeted Visual Reasoning Data via Synthetic Augmentation}
To further strengthen visual reasoning after the first stage, we construct a targeted multimodal dataset by employing a synthetic data generation pipeline to large collections of raw images. The pipeline transforms each image into one or more task-centric training samples. This shifts the original image distribution to a custom curriculum that encourages the model to learn spatial structure, compositionality, and fine-grained perception that transfer to more complex visual reasoning. The following are the primary categories we target:

\begin{itemize} [noitemsep, topsep=0pt, leftmargin=*]

  \item \textbf{Image Reconstruction}: Learn holistic scene priors and part–whole reasoning by masking image regions.

  \item \textbf{Visual Matching}: Improve correspondence, retrieval, and fine-grained discrimination by matching cropped or augmented anchors to candidates across views or images.

  \item \textbf{Object Detection}: Strengthen grounding and localization by identifying object presence and approximate location.

  \item \textbf{Counting}: Enhance the ability to count and distinguish specific visual elements by querying total or category-specific counts.

\end{itemize}

\paragraph{Data Hygiene and Difficulty Control}
For each task, we modulate difficulty through controlled augmentation depending on the task. This helps the model learn a more robust spatial reasoning, compositional understanding, and precise grounding, while remaining broadly applicable across diverse visual domains.

\paragraph{Training Setup}
In this stage, the vision encoder was frozen, with just the projection network and decoder updated during training. The training was performed at a sequence length of 16384 (with sequence packing), and learning rate of 1e-5 with cosine decay and 10\% warmup. For samples having an instruction-response format, we compute loss only on the reponses in this stage.  The final checkpoint from this stage was considered as the base model for future stages.

\paragraph{Evaluating effectiveness of Stage-2} To evaluate the effectiveness of the second CPT stage, we conducted two small-scale SFT experiments, initialized from the final checkpoints of CPT Stage 1 and Stage 2, using 17k text-based reasoning samples designed to mimic our full SFT setup. 
Table \ref{tab:cpt_stage_eval} presents a comparative evaluation of SFT after the two CPT stages across a range of multimodal and math-focused vision benchmarks (see \ref{sec:vision_benchmarks}). Stage 2 consistently improves performance over Stage 1, with notable gains on tasks such as MathVerse (Vision Dominant: +9.65 points), CharXiv (Descriptive: +5.98 points), and AI2D Test (+3.7 points). These results demonstrate that CPT Stage 2 provides substantial benefits for visual reasoning tasks.

\begin{table}[ht]
\centering
\footnotesize
\begin{tabular}{@{}lcc@{}}
\toprule
\textbf{Benchmark} & 
\textbf{SFT on CPT Stage 1} & 
\textbf{SFT on CPT Stage 2} \\
\midrule
MMMU (Val) & 64.11 & 69.10 \\

MathVision & 44.4 & 47.36 \\

MathVista & 71.8 & 74.10 \\

MathVerse (Vision Dominant) & 53.04 & 62.69 \\

MathVerse (Text Dominant) & 70.81 & 78.42 \\

MMStar & 61.80 & 66.30 \\

CharXiv (descriptive) & 80.22 & 86.20 \\

CharXiv (reasoning) & 43.5 & 48.00 \\

AI2D Test & 78.1 & 81.8 \\
\bottomrule
\end{tabular}
\vspace{+2mm}
\caption{Evaluating effectiveness of CPT Stage 2 across multiple vision benchmarks.}
\label{tab:cpt_stage_eval}
\end{table}


\section{Supervised Fine Tuning (SFT)} \label{sec:sft}

Following the upscaling and continual pretraining stages, which yielded a base model with strong reasoning capabilities, we performed Supervised Fine-Tuning (SFT) to develop the model into a full-fledged reasoner.

\paragraph{Data Curation}
 Given compute constraints that preclude training a larger annotator model or scaling post-training runs from a cold-start SFT, we emphasize curating and synthesizing high-quality, high-signal prompts and employ open-source models as annotators. We curate and synthesize a diverse set of prompts\cite{slam-distillation-from-r1}. Small-scale ablations using DeepSeek-R1-0528 \cite{deepseekai2025deepseekr1incentivizingreasoningcapability} and gpt-oss-120b \cite{openai2025gptoss120bgptoss20bmodel} presented in Table \ref{tab:annotator_comparison} show minimal performance differences between annotators for our base model. We therefore adopt gpt-oss-120b as our annotator model due to its greater compute efficiency. For verifiable domains, such as Math, Coding and Science, we follow the synthetic data generation methodology in \cite{tiwari-etal-2025-auto} to synthesize high quality, execution verifiable data samples across domains starting from a seed taxonomy and samples, and evolving  iteratively toward more complex scenarios \cite{xu2025wizardlmempoweringlargepretrained}. That said, this release prioritized performance, with safety mitigations included but not pursued to the same depth.
 
 To ensure the highest data quality and maximize sample efficiency, we invested significantly in a comprehensive data processing pipeline. We followed a multi-step filtering process that included rigorous de-duplication to enhance data diversity, content filtering to remove unsafe or inappropriate material, and heuristic filtering to remove low-quality samples. Following this initial cleaning, we verified the data's correctness using  LLM-as-Judge and execution-based verification where applicable, implementing rejection sampling to discard incorrect or low-quality instruction-response pairs. This verification stage also included format-based checks to ensure structural correctness for samples where a specific output like JSON or XML was expected. Finally, all samples were processed with consistent formatting using our custom chat template, and a decontamination stage to remove any samples overlapping with the benchmarks.

\begin{table}[ht]
\centering
\footnotesize
\begin{tabular}{lcc@{}}
\toprule
\textbf{Benchmark} & \textbf{Annotator: DeepSeek-R1-0528} & \textbf{Annotator: gpt-oss-120b}  \\
\midrule
GPQA Diamond & 67.67 & 65.82 \\
AIME'24      & 80.66 & 80.67 \\
AIME'25      & 75.33 & 74 \\
\bottomrule
\end{tabular}
\vspace{+2mm}
\caption{Performance on benchmarks relevant to a small scale SFT set, annotated with DeepSeek-R1-0528 and gpt-oss-120b. Overall, we find the benchmark performance to be similar for both annotator models.}
\label{tab:annotator_comparison}
\end{table}

\paragraph{Data Composition}
We use a large and diverse dataset containing millions of high-quality instruction–response pairs. Each response contained explicit reasoning steps leading to the final response, followed by the final response itself. 
The final dataset comprised samples from domains including mathematical reasoning, coding,  scientific reasoning, tool calling, generic reasoning and knowledge-intensive samples, conversations, instruction-following, security, content moderation, and robustness. This ensures that the model is both capable and reliable across diverse scenarios.

\paragraph{Training}
 We first performed an initial SFT for 4 epochs at a sequence length of 32768 (with sequence packing) and a learning rate of 1e-5 with cosine decay. To further improve performance, we conducted two smaller SFT runs on top of the large-scale SFT: (1) trained with a stratified 25\% subset of the full dataset for 4 more epochs at the same sequence length and (2) a longer-sequence run at 49,152 sequence length, using 25k samples between 32768 and 49152 tokens and 100k samples $\leq$ 32768 tokens, randomly drawn from the original mix. The models from these two smaller runs were merged by averaging their weights in equal proportions to produce the final \textsc{Apriel-}1.5-15B-\textsc{Thinker} checkpoint. These smaller runs provided inexpensive gains in overall and long-context performance, and the merge balanced the benefits of both. As this phase consisted entirely of text data, only the decoder was updated. In all SFT runs, loss was computed only on response, and the chat template was applied to all samples.  

\section{Evaluation Methodology} \label{sec:eval}

\subsection{Text evaluation}

To report the evaluation results for the \modelname, we relied on the \textbf{Artificial Analysis Intelligence Index}, 
an independent combination metric for measuring general intelligence in large language models (LLMs). Using this external source ensures that results are unbiased and comparable across organizations, as the scoring is not influenced by in-house test sets or proprietary metrics. Although our internal evaluation also show very similar metrics as reported by Artificial Intelligence. 

The Artificial Analysis Intelligence Index is notable for its \emph{breadth and methodological rigor}. 
It aggregates results from ten heterogeneous benchmarks, with each benchmark targeting a distinct dimension of model capability:

\begin{itemize} [leftmargin=*]
    \item \textbf{MMLU-Pro} -- advanced multi-domain knowledge and reasoning
    \item \textbf{GPQA Diamond} -- graduate-level problem solving in science/engineering
    \item \textbf{Humanity’s Last Exam} -- multi-disciplinary high-difficulty reasoning
    \item \textbf{LiveCodeBench} -- functional correctness in code generation
    \item \textbf{SciCode} -- scientific computing and reasoning tasks
    \item \textbf{AIME 2025} -- competition-level mathematics
    \item \textbf{IFBench} -- instruction following and compliance
    \item \textbf{AA-LCR} -- long-context reasoning
    \item \textbf{Terminal-Bench Hard} -- real-world Linux shell execution and system tool use in end-to-end tasks
    \item \textbf{$\tau^2$-Bench Telecom} -- specialized domain evaluation in applied tasks
\end{itemize}

By normalizing across domains, evaluation difficulty, and inter-benchmark variance, the Index 
provides a \emph{holistic measure of intelligence} rather than domain-specific performance. 
This methodology makes it a well-respected yardstick across academia and industry\footnote{https://artificialanalysis.ai/methodology/intelligence-benchmarking}.

\subsection{Vision evaluation}\label{sec:vision_benchmarks}

For the vision component, we focus on image evaluations since our training has been conducted primarily with images. We evaluate vision capabilities using the VLMEvalKit\cite{duan2024vlmevalkit} toolkit, which standardizes data loading, prompting, post-processing, and scoring for reproducible comparisons across diverse tasks. Our benchmark suite spans the following areas:
\begin{itemize} [noitemsep, topsep=0pt, leftmargin=*]

  \item \textbf{General Multi-modal Reasoning}
    \begin{itemize}
        \item MMMU\cite{yue2023mmmu}: Multi-modal understanding benchmark focusing on evaluating visual knowledge and reasoning.
        \item MMMU-Pro\cite{yue2024mmmu}: Enhanced Multi-modal understanding benchmark focusing on evaluating visual knowledge and reasoning.
        \item MMStar\cite{chen2024mmstar}: Vision-indispensable benchmark focusing on tasks that cannot be solved with only knowledge or without using the image.
    \end{itemize}
  \item \textbf{Visual Logic} 
    \begin{itemize}
        \item LogicVista\cite{xiao2024logicvista}: Multi-modal logical reasoning benchmark targeting different reasoning skill types in visual contexts.
    \end{itemize}
  \item \textbf{Mathematical Vision and Quantitative Reasoning} 
    \begin{itemize}
        \item MathVision\cite{wang2024mathvision}: Mathematical reasoning within visual contexts.
        \item MathVista\cite{lu2023mathvista}: Benchmark combining challenges from various visual and mathematical tasks.
        \item MathVerse\cite{zhang2024mathverse}: Mathematical benchmark measuring model performance across different levels of information content across multiple modalities.
    \end{itemize}
  \item \textbf{Document/Diagram Understanding} 
    \begin{itemize}
        \item CharXiv\cite{wang2024charxiv}: Benchmark measuring descriptive and reasoning question answering capabilities across basic and complex chart elements respectively.
        \item AI2D\cite{kembhavi2016ai2d}: Diagram understanding benchmark.
    \end{itemize}
  \item \textbf{Open-domain Vision–Language Reasoning} 
    \begin{itemize}
        \item BLINK\cite{fu2024blink}: Benchmark measuring performance on various visual perception tasks.
    \end{itemize}

\end{itemize}
For each dataset, we adhere to official or community-standard protocols as implemented in VLMEvalKit and adopt consistent prompts and inference settings across models to ensure fair comparisons.

\section{Results and Observations}
\label{Benchmarking}

\subsection{Text Benchmarks} 

\begin{figure}[htbp]
    \centering
    \captionsetup{font=small}
    \begin{subfigure}[b]{\textwidth}
        \centering
        \includegraphics[width=\linewidth]{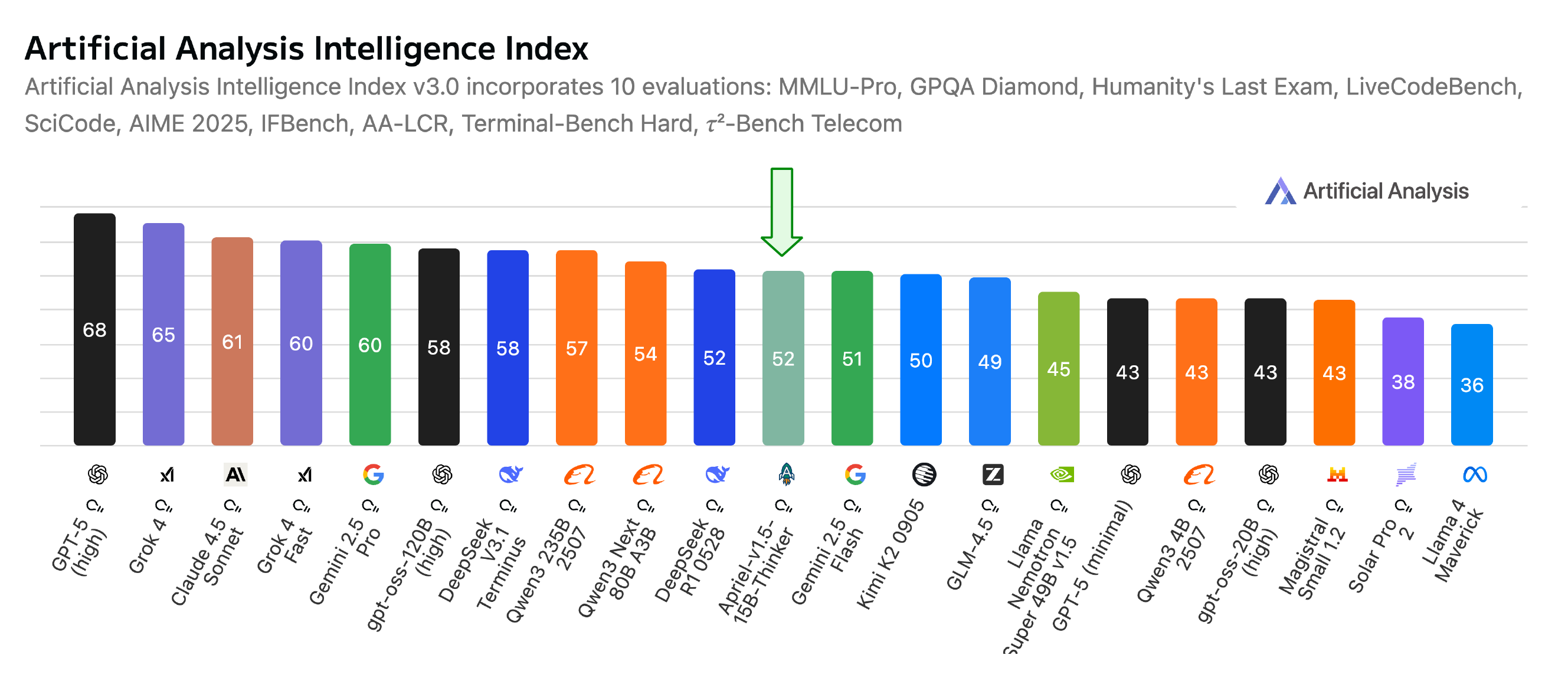}
        \caption{\modelname\ compared with state-of-the-art LLMs.}
    \end{subfigure}
    \vskip\baselineskip
    \begin{subfigure}[b]{\textwidth}
        \centering
        \includegraphics[width=\linewidth]{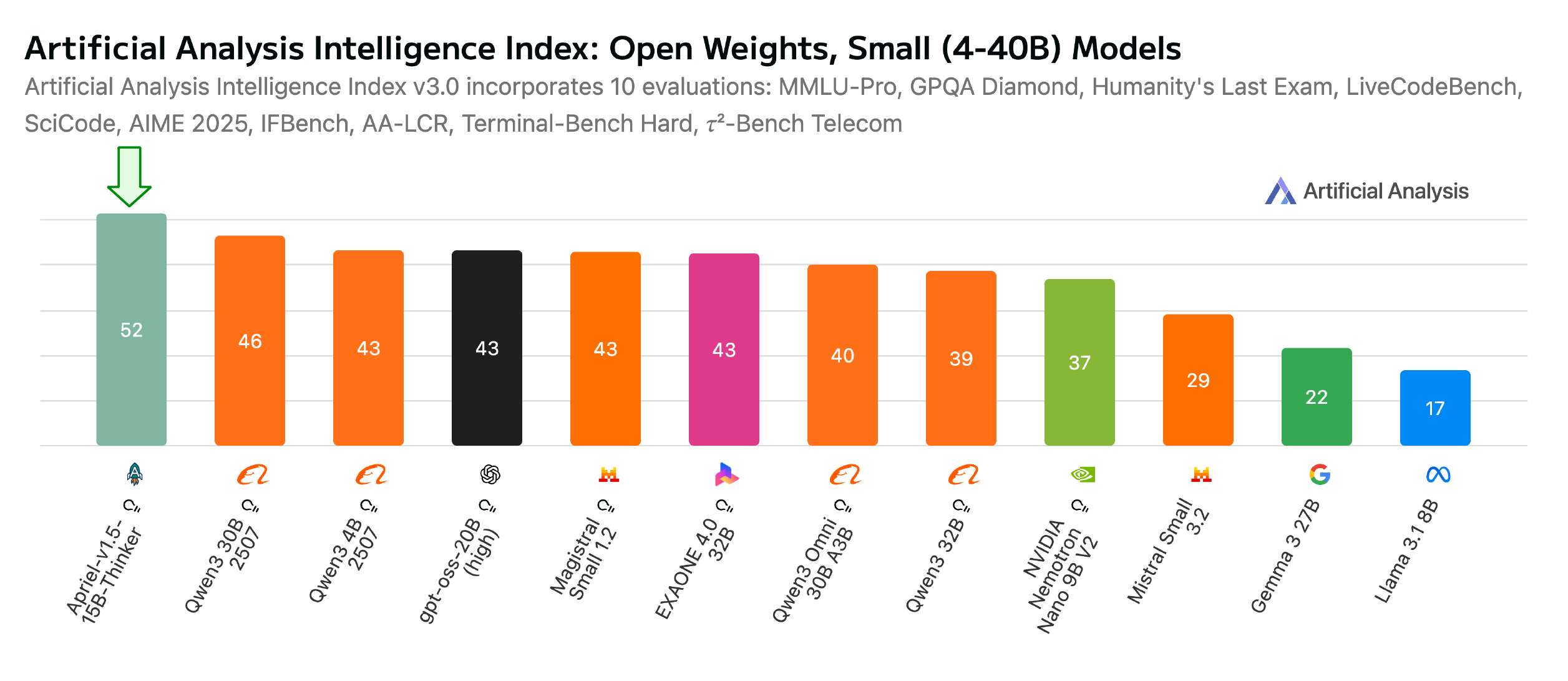}
        \caption{\modelname\ compared with SOTA open-source models.}
    \end{subfigure}
    \caption{\modelname\ ranks first in Artificial Analysis Intelligence index among the SOTA small open-source models and delivers performance competitive to larger open-source and proprietary models (as of September 26th, 2025).}
    \label{fig:main-results}
\end{figure}

\begin{figure}[!htbp]
    \centering
    \captionsetup{font=small}
    \includegraphics[width=1.0\textwidth]{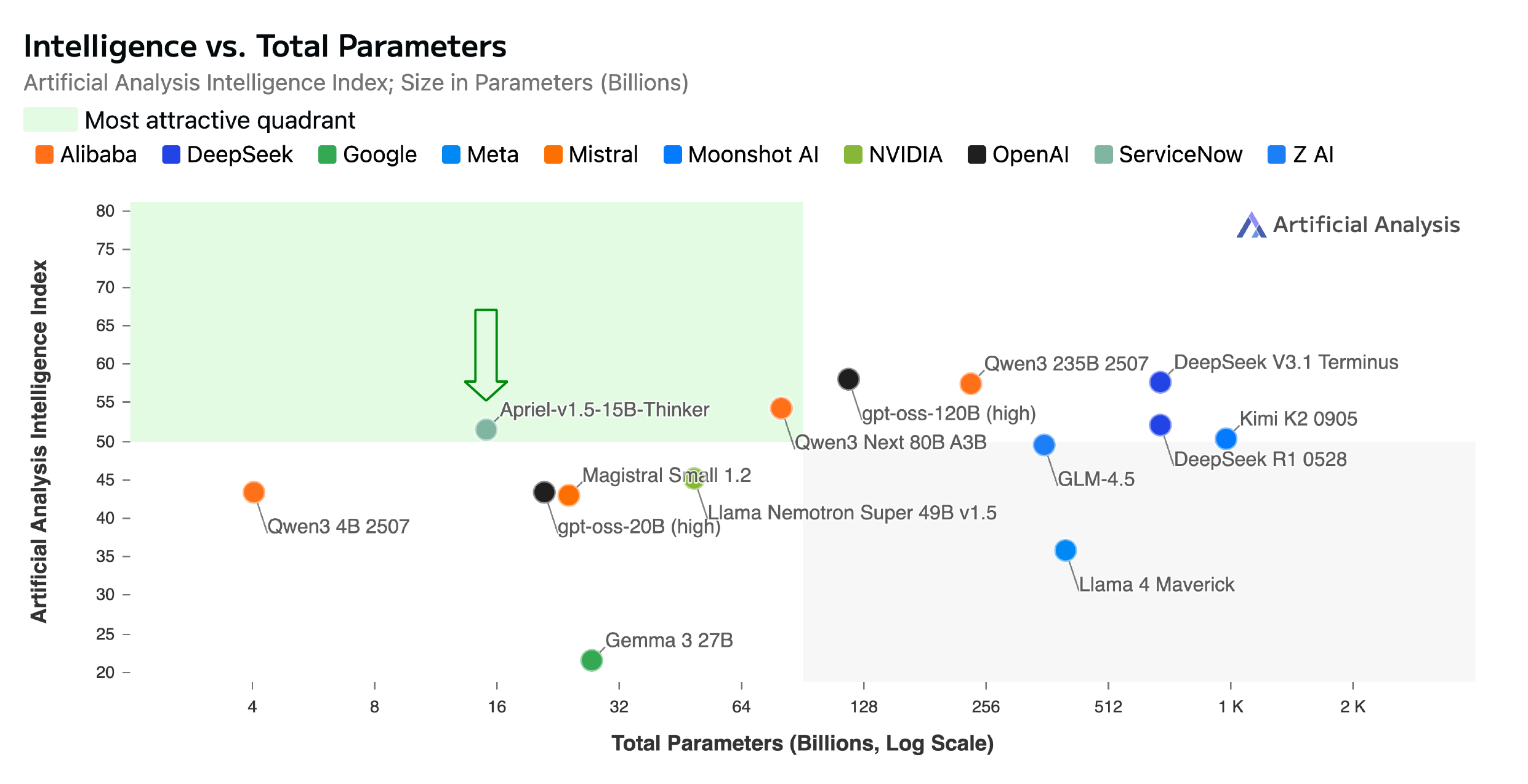}
    \caption{Artificial Analysis Intelligence Index vs. Total Parameters (log scale). 
    \modelname\ lies in the ``most attractive quadrant.''}
    \label{fig:aa-params}
\end{figure}

Figure~\ref{fig:main-results} shows \modelname\ achieves a score of \textbf{52}. It surpasses larger open-weight systems such as \textbf{Llama Nemotron Super 49B v1.5 (45)} and \textbf{gpt-oss-20B (43)}, while performing comparably to models such as \textbf{DeepSeek-R1-0528} and \textbf{Gemini-2.5-Flash}.


The aggregated results across the AA intelligence index offer a consolidated view of the model’s reasoning performance. The model demonstrates state-of-the-art accuracy on several challenging evaluations, achieving 87\% on AIME2025, 62\% on IF-Bench, and 68\% on $\tau^2$Bench (Telecom). These results highlight its strong mathematical reasoning, robust instruction-following, and domain-specific problem-solving capabilities, where it consistently outperforms significantly larger open-source baselines.

On TerminalBench-Hard, the model achieves a score of 10\%. This benchmark evaluates performance in terminal-based environments across a wide spectrum of technical tasks. Despite its smaller parameter count, our model performs competitively with much larger proprietary systems such as GPT-4.1 and Gemini 2.5 Flash (both at 13\%) and Qwen3-250B (13\%). Notably, it surpasses strong open-source peers of comparable size, including \texttt{gpt-oss-20b}, which scores 6\%.

These results underscore the efficiency and competitiveness of the model, offering strong reasoning and agentic capabilities without the overhead of massive parameter counts.

The detailed breakdown in Table~\ref{tab:custom-benchmark-results} reports the scores provided by \emph{Artificial Analysis}, with the exception of \modelname\ (self-reported), where evaluations were conducted internally. 

The divergence in internal evaluation arises primarily from differences in benchmarking conditions. In particular, the judging models for AA-LCR and $\tau^2$Bench differ (\texttt{GPT-4.1} versus \texttt{Qwen3-235B-A22B-2507}). For \texttt{AIME2025}, no language model equality checker was employed. Moreover, \texttt{TerminalBench} was executed under a shorter timeout constraint, further contributing to lower scores.

\begin{table}[htbp!]
\centering
\captionsetup{font=small}
\resizebox{\columnwidth}{!}{%
\begin{tabular}{lccccccccccc}
\toprule
\textbf{Benchmark} 
 & \textbf{\thead{Artificial Analysis\\ Intelligence Index}}
 & \textbf{MMLU-Pro} 
 & \textbf{GPQA Diamond} 
 & \textbf{HLE} 
 & \textbf{LiveCodeBench} 
 & \textbf{SciCode} 
 & \textbf{AIME2025} 
 & \textbf{IF-Bench} 
 & \textbf{AA-LCR} 
 & \textbf{\thead{TerminalBench\\ Hard}} 
 & \textbf{$\tau^2$-Bench Telecom} \\
\midrule
\rowcolor{gray!10} \multicolumn{12}{c}{Proprietary Models} \\ \midrule
\rowcolor{orange!10} GPT-5 (High) & 68.47 & 87.1 & 85.4 & 26.5 & 84.6 & 42.9 & 94.3 & 73.1 & 75.6 & 30.5 & 84.8 \\
\rowcolor{orange!10} Grok 4 & 65.26 & 86.6 & 87.7 & 23.9 & 81.9 & 45.7 & 92.7 & 53.7 & 68 & 37.6 & 74.9 \\
\rowcolor{orange!10} Claude 4.5 Sonnet & 61.29 & 87.5 & 83.4 & 17.3 & 57.7 & 44.7 & 88 & 57.3 & 65.7 & 33.3 & 78.1 \\
\rowcolor{orange!10} Grok 4 Fast & 60.25 & 85 & 84.7 & 17 & 83.2 & 44.2 & 89.7 & 50.5 & 64.7 & 17.7 & 65.8 \\
\rowcolor{orange!10} Gemini 2.5 Pro & 59.59 & 86.2 & 84.4 & 21.1 & 80.1 & 42.8 & 87.7 & 48.7 & 66 & 24.8 & 54.1 \\
\rowcolor{orange!10} Claude 4.1 Opus & 59.27 & 88 & 80.9 & 11.9 & 65.4 & 40.9 & 80.3 & 55.4 & 66.3 & 32.1 & 71.4 \\
\rowcolor{orange!10} Claude 4 Sonnet & 56.51 & 84.2 & 77.7 & 9.6 & 65.5 & 40 & 74.3 & 54.7 & 64.7 & 29.8 & 64.6 \\
\rowcolor{orange!10} Magistral Medium 1.2 & 52.05 & 81.5 & 73.9 & 9.6 & 75 & 39.2 & 82 & 43 & 51.3 & 12.8 & 52 \\
\rowcolor{orange!10} Gemini 2.5 Flash & 51.18 & 83.2 & 79 & 11.1 & 69.5 & 39.4 & 73.3 & 50.3 & 61.7 & 12.8 & 31.6 \\
\rowcolor{orange!10} GPT-5 (Minimal) & 43.48 & 80.6 & 67.3 & 5.4 & 55.8 & 38.8 & 31.7 & 45.6 & 25 & 17.7 & 67 \\

\midrule
\rowcolor{gray!10} \multicolumn{12}{c}{Large Open Weight Models} \\ \midrule
\rowcolor{blue!10} gpt-oss-120B (High) & 57.98 & 80.8 & 78.2 & 18.5 & 65.3 & 36.2 & 93.4 & 69 & 50.7 & 22 & 65.8 \\
\rowcolor{blue!10} DeepSeek v3.1 Terminus & 57.71 & 85.1 & 79.2 & 15.2 & 79.8 & 40.6 & 89.7 & 57 & 65 & 28.4 & 37.1 \\
\rowcolor{blue!10} Qwen3 235B 2507 & 57.47 & 84.3 & 79 & 15 & 78.8 & 42.4 & 91 & 51.2 & 67 & 12.8 & 53.2 \\
\rowcolor{blue!10} DeepSeekR1 0528 & 52.01 & 84.9 & 81.3 & 14.9 & 77 & 40.3 & 76 & 39.6 & 54.7 & 14.9 & 36.5 \\
\rowcolor{blue!10} Kimi K2 0905 & 50.4 & 81.9 & 76.7 & 6.3 & 61 & 30.7 & 57.3 & 41.7 & 52.3 & 22.7 & 73.4 \\
\rowcolor{blue!10} GLM 4.5 & 49.44 & 83.5 & 78.2 & 12.2 & 73.8 & 34.8 & 73.7 & 44.1 & 48.3 & 21.3 & 24.6 \\
\rowcolor{blue!10} GLM 4.5 Air & 48.81 & 81.5 & 73.3 & 6.8 & 68.4 & 30.6 & 80.7 & 37.6 & 43.7 & 19.1 & 46.5 \\
\rowcolor{blue!10} MiniMax M1 80k & 46.22 & 81.6 & 69.7 & 8.2 & 71.1 & 37.4 & 61 & 41.8 & 54.3 & 2.8 & 34.2 \\
\rowcolor{blue!10} Llama 4 Maverick & 35.8 & 80.9 & 67.1 & 4.8 & 39.7 & 33.1 & 19.3 & 43 & 46 & 6.4 & 17.8 \\
\rowcolor{blue!10} Llama 4 Scout & 28.14 & 75.2 & 58.7 & 4.3 & 29.9 & 17 & 14 & 39.5 & 25.8 & 1.4 & 15.5 \\

\midrule
\rowcolor{gray!10} \multicolumn{12}{c}{Small Open Weight Models} \\ \midrule
\rowcolor{green!10} Qwen3 Next 80B A3B & 54.32 & 82.4 & 75.9 & 11.7 & 78.4 & 38.8 & 84.3 & 60.7 & 60.3 & 9.2 & 41.5 \\
\rowcolor{green!10} Qwen3 30B 2507 & 46.41 & 80.5 & 70.7 & 9.8 & 70.7 & 33.3 & 56.3 & 50.7 & 59 & 5 & 28.1 \\
\rowcolor{green!10} gpt-oss-20B (High) & 43.27 & 73.6 & 61.7 & 8.5 & 57.2 & 35.4 & 61.7 & 60.5 & 18.7 & 5.7 & 49.7 \\
\rowcolor{green!10} Llama Nemotron Super 49B v1.5 & 45.22 & 81.4 & 74.8 & 6.8 & 73.7 & 34.8 & 76.7 & 37 & 34 & 5 & 28.1 \\
\rowcolor{green!10} Qwen3 4B 2507 & 43.36 & 74.3 & 66.7 & 5.9 & 64.1 & 25.6 & 82.7 & 49.8 & 37.7 & 1.4 & 25.4 \\
\rowcolor{green!10} Magistral Small 1.2 & 42.97 & 76.8 & 66.3 & 6.1 & 72.3 & 35.2 & 80.3 & 44.4 & 16.3 & 4.3 & 27.8 \\
\rowcolor{green!10} exaone 4.0 32B & 42.64 & 81.8 & 73.9 & 10.5 & 74.7 & 34.4 & 80 & 36.3 & 14 & 3.5 & 17.3 \\
\rowcolor{green!10} Nvidia Nemotron Nano 9B V2 & 36.91 & 74.2 & 57 & 4.6 & 72.4 & 22 & 69.7 & 27.6 & 18.3 & 1.4 & 21.9 \\
\rowcolor{green!10} Mistral Small 3.2 & 29.05 & 68.1 & 50.5 & 4.3 & 27.5 & 26.5 & 27 & 33.5 & 17.3 & 6.4 & 29.5 \\
\rowcolor{green!10} Llama 3.1 8B & 16.91 & 47.6 & 25.9 & 5.1 & 11.6 & 13.2 & 4.3 & 28.6 & 15.7 & 0.7 & 16.4 \\
\midrule
\rowcolor{green!20} \textbf{\modelname} & 51.57 & 77.3 & 71.3 & 12 & 72.8 & 34.8 & 87.5 & 61.7 & 20 & 9.9 & 68.4 \\
\rowcolor{green!20} \textbf{\modelname\ (self-reported)} & 50 & 76.48 & 70.61 & 11 & 71.6 & 36.46 & 83.67 & 60.45 & 26.33 & 5.7 & 57.8 \\

\bottomrule
\end{tabular}%
}
\vspace{+2mm}
\caption{Evaluation (pass@1 or accuracy) on benchmarks with \textbf{maximum reasoning}, as applicable: MMLU-Pro, GPQA Diamond, HLE, LiveCodeBench, SciCode, AIME2025, IF-Bench, AA-LCR, TerminalBench-Hard, and $\tau^2$-2Bench. Orange = proprietary models, blue = $>$50B open-weight models, green = $<$50B open-weight models.}
\label{tab:custom-benchmark-results}
\end{table}

Figure~\ref{fig:aa-params} demonstrates performance relative to scale, and falls within the \emph{``most attractive quadrant''} -- the region where models combine moderate scale with disproportionately high performance. This placement underscores \modelname’s \emph{superior cost-to-intelligence trade-off}, 
offering reduced compute requirements and faster inference while maintaining robust 
general capabilities. The findings support the broader trend that 
\emph{smaller, efficiently trained models can close the gap with frontier models}.

\subsection{Vision Benchmarks} 

\begin{figure}[t]
  \centering
  \captionsetup{font=small}
  \includegraphics[width=0.75\linewidth]{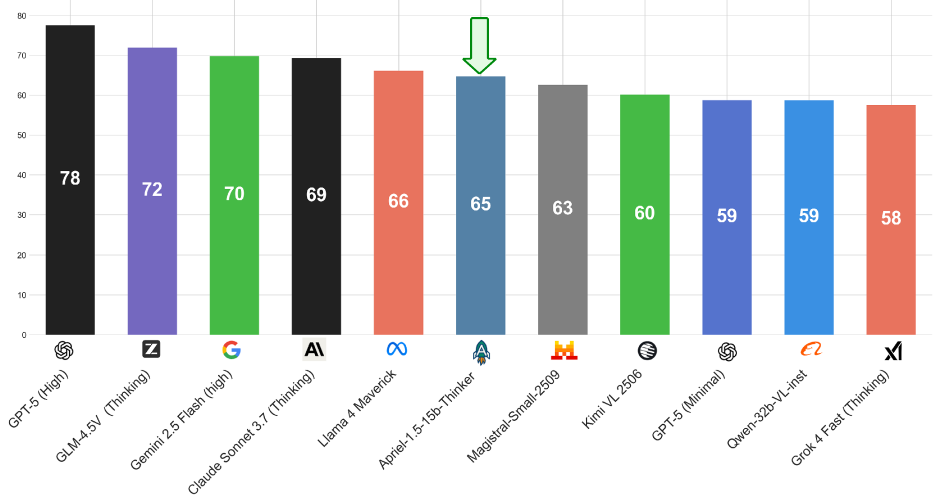}
  \caption{Average performance across the benchmark suite (higher is better). The chart aggregates scores from MMMU~\cite{yue2023mmmu}, MMMU-Pro~\cite{yue2024mmmupro}, LogicVista~\cite{xiao2024logicvista}, MathVision~\cite{wang2024mathvision}, MathVista~\cite{lu2023mathvista}, MathVerse~\cite{zhang2024mathverse}, MMStar~\cite{chen2024mmstar}, CharXiv~\cite{wang2024charxiv}, AI2D~\cite{kembhavi2016ai2d}, and BLINK~\cite{fu2024blink}.}
  \label{fig:vision-avg}
\end{figure}

Figure~\ref{fig:vision-avg} summarizes model-level averages over the full image benchmark suite, highlighting overall capability and robustness. Detailed per-benchmark, per-model scores are reported in Table~\ref{tab:vlm-benchmark-results}, where higher values indicate better performance. Together, these views provide both a high-level comparison and fine-grained insight into strengths and weaknesses across categories.

\begin{table}[htbp!]
\centering
\captionsetup{font=small}
\resizebox{\columnwidth}{!}{%
\begin{tabular}{lcccccccccccccc}
\toprule
\multicolumn{1}{l}{\multirow{2}{*}{\textbf{Benchmark}}} & \multirow{2}{*}{\textbf{MMMU}} & \multicolumn{2}{c}{\textbf{MMMU-PRO}} & \multirow{2}{*}{\textbf{LogicVista}} & \multirow{2}{*}{\textbf{MathVision}} & \multirow{2}{*}{\textbf{MathVista}} & \multicolumn{2}{c}{\textbf{MathVerse}} & \multirow{2}{*}{\textbf{MMStar}} & \multicolumn{2}{c}{\textbf{CharXiv}} & \textbf{AI2D} & \multirow{2}{*}{\textbf{BLINK}} & \multirow{2}{*}{\textbf{Average}} \\ \cmidrule(lr){3-4} \cmidrule(lr){8-9} \cmidrule(lr){11-13}
\multicolumn{1}{c}{} & \textbf{Val} & \textbf{10 C} & \textbf{vision} &  &  &  & \textbf{Vision-dom} & \textbf{Text-dom} &  & \textbf{Des} & \textbf{R} & \textbf{test} &  &  \\ \midrule
\rowcolor{gray!10} \multicolumn{15}{c}{Proprietary Models} \\ \midrule
\rowcolor{orange!10} GPT-5 (High) & 81.33 & 74.73 & 66.93 & 69.35 & 67.10 & 83.30 & 79.82 & 84.64 & 77.74 & 91.25 & 71.50 & 90.05 & 70.22 & 77.54 \\
\rowcolor{orange!10} GPT-5 (Minimal) & 66.66 & 66.06 & 57.68 & 44.51 & 35.52 & 61.20 & 39.84 & 43.78 & 63.60 & 82.45 & 52.80 & 85.16 & 64.59 & 58.76 \\
\rowcolor{orange!10} Grok 4 Fast (Thinking) & 70.11 & 61.61 & 22.94 & 47.42 & 48.35 & 68.20 & 54.69 & 72.20 & 64.80 & 68.15 & 33.50 & 81.86 & 54.39 & 57.56 \\
\rowcolor{orange!10} Claude Sonnet 3.7 (Thinking) & 73.66 & 64.5 & 60.11 & 69.12 & 50.32 & 74.60 & 56.09 & 69.28 & 70.00 & 93.27 & 70.90 & 84.19 & 64.49 & 69.27 \\
\rowcolor{orange!10} Gemini 2.5 Flash (High) & 70.66 & 67.86 & 56.76 & 63.75 & 59.21 & 78.50 & 70.68 & 78.80 & 73.86 & 83.60 & 56.50 & 82.09 & 65.64 & 69.84 \\ \midrule
\rowcolor{gray!10} \multicolumn{15}{c}{Large Open Weight Models} \\ \midrule
\rowcolor{blue!10} Llama 4 Maverick & 72.22 & 63.41 & 54.45 & 58.38 & 43.09 & 72.60 & 63.32 & 70.30 & 71.00 & 87.70 & 55.00 & 85.78 & 62.28 & 66.12 \\
\rowcolor{blue!10} GLM-4.5V (Thinking) & 74.33 & 64.16 & 61.50 & 63.53 & 59.53 & 83.60 & 68.65 & 77.41 & 74.46 & 90.80 & 63.00 & 87.75 & 66.59 & 71.95 \\
\midrule
\rowcolor{gray!10} \multicolumn{15}{c}{Small Open Weight Models} \\ \midrule\rowcolor{green!10} Qwen 2.5 VL 32B-Instruct & 64.40 & 51.90 & 45.02 & 52.15 & 38.15 & 75.90 & 51.64 & 62.69 & 66.30 & 72.50 & 39.10 & 81.60 & 61.90 & 58.71 \\
\rowcolor{green!10} Magistral Small-2509 & 70.00 & 55.72 & 46.06 & 54.80 & 55.92 & 73.40 & 53.68 & 67.76 & 65.13 & 82.27 & 52.90 & 82.18 & 54.48 & 62.64 \\
\rowcolor{green!10} Kimi VL 2506 & 61.44 & 48.09 & 42.89 & 46.97 & 50.00 & 79.80 & 52.91 & 67.13 & 68.93 & 78.70 & 45.60 & 82.70 & 56.33 & 60.11 \\
\midrule
\rowcolor{green!20} \textbf{\modelname} & 70.22 & 55.38 & 48.21 & 58.39 & 50.99 & 75.50 & 58.38 & 76.40 & 67.73 & 88.20 & 50.10 & 82.87 & 58.71 & 64.70 \\ \bottomrule
\end{tabular}%
}
    \vspace{+2mm}
    \caption{Evaluation (pass@1 or accuracy, as applicable) on multimodal benchmarks covering general reasoning (MMMU, MMStar), visual logic (LogicVista), mathematical vision tasks (MathVision, MathVista, MathVerse), document-level understanding (CharXiv), diagram understanding (AI2D), and open-domain vision-language reasoning (BLINK). orange = proprietary models, blue = >100B open weight models, green = <50B open weight models.}
  \label{tab:vlm-benchmark-results}
\end{table}

 As shown in Figure~\ref{fig:vision-avg}, the strongest overall performance is achieved by larger, advanced reasoning models, with notable leads on broad multimodal and STEM-oriented tasks. \emph{\modelname} attains a solid position among the evaluated models, beating most similarly-sized and even larger open-weights vision-language models such as Kimi-VL-2506\cite{kimiteam2025kimivltechnicalreport} and Qwen-2.5-VL-3B-Instruct\cite{bai2025qwen25vl}. Despite its smaller size (15B parameters), Apriel closely tracks much larger models such as Llama~4 Maverick\cite{llama4} (400B parameters) and outperforms several larger proprietary baselines (e.g., GPT-5~Minimal~\cite{openai2025gpt5systemcard}, Grok~4~Fast~\cite{xai2025grok4fastsystemcard}) in the overall average score. 

The detailed breakdown in Table~\ref{tab:vlm-benchmark-results} indicates strong results on document-centric and diagram understanding benchmarks (e.g., CharXiv, AI2D), competitive performance on general multimodal reasoning (MMStar), and visual mathematical skills (MathVista). Apriel demonstrates particularly strong performance in document understanding tasks, achieving 88.20\% on CharXiv descriptive tasks, which is the third-highest score after Claude and GPT-5 (High). Similarly, on MathVerse (Text Dominant), Apriel scores 76.40\%, outperforming several larger models including Claude Sonnet, Magistral, and Llama 4 Maverick\cite{llama4}.
The results suggest a pattern where Apriel performs better on tasks that combine visual inputs with substantial textual reasoning components, while showing moderate performance on purely visual reasoning tasks. For instance, on MMMU (70.22\%), Apriel demonstrates competitive performance, whereas on vision-dominant tasks like MMMU-PRO (Vision) at 48.21\%, it shows room for improvement.
Most models, including Apriel, show stronger performance on structural understanding tasks (AI2D) and descriptive document tasks (CharXiv descriptive) compared to more complex reasoning tasks (CharXiv reasoning, LogicVista). The Apriel model demonstrates this pattern as well, with a notable 38.1 percentage point difference between its performance on CharXiv descriptive (88.20\%) and CharXiv reasoning (50.10\%) tasks, highlighting a gap between surface-level document comprehension and deeper contextual reasoning. Performance on the most demanding STEM-centric and visual logic tasks remains a key opportunity for further improvement.


\section{Conclusion and Future Work} \label{sec:conclusion}
Our work shows that a 15-billion-parameter model can reach frontier-level reasoning by prioritizing data quality and a deliberately structured \textbf{mid-training} pipeline—staged continual pretraining (CPT) followed by large-scale, high-signal SFT without reinforcement learning or preference optimization.   This data-centric recipe yields measurable gains during CPT (e.g., \textbf{+9.65} on MathVerse Vision-Dominant) and culminates in strong text-reasoning results on \textbf{AIME} and \textbf{GPQA}, while remaining competitive on multimodal benchmarks.    Crucially, the final model operates on a \textbf{single-GPU}, delivering a favorable performance–efficiency trade-off that makes frontier-level reasoning accessible to organizations with limited computational infrastructure.


While this work focused primarily on text-based reasoning, the model’s multimodal results offer a solid foundation for future development. Our next steps will extend multimodal capabilities more comprehensively and strengthen agentic abilities to support interactive workflows, with targeted alignment techniques where appropriate. Future development will continue to be guided by the core principles demonstrated here: strategic mid-training design,  efficient architectural scaling and a continued focus on high-quality, targeted data.

\printbibliography

@misc{radhakrishna2025aprielnemotron15bthinker,
      title={Apriel-Nemotron-15B-Thinker}, 
      author={Shruthan Radhakrishna and Soham Parikh and Gopal Sarda and Anil Turkkan and Quaizar Vohra and Raymond Li and Dhruv Jhamb and Kelechi Ogueji and Aanjaneya Shukla and Oluwanifemi Bamgbose and Toby Liang and Luke Kumar and Oleksiy Ostapenko and Shiva Krishna Reddy Malay and Aman Tiwari and Tara Bogavelli and Vikas Yadav and Jash Mehta and Saloni Mittal and Akshay Kalkunte and Pulkit Pattnaik and Khalil Slimi and Anirudh Sreeram and Jishnu Nair and Akintunde Oladipo and Shashank Maiya and Khyati Mahajan and Rishabh Maheshwary and Masoud Hashemi and Sai Rajeswar Mudumba and Sathwik Tejaswi Madhusudhan and Torsten Scholak and Sebastien Paquet and Sagar Davasam and Srinivas Sunkara},
      year={2025},
      eprint={2508.10948},
      archivePrefix={arXiv},
      primaryClass={cs.LG},
      url={https://arxiv.org/abs/2508.10948}, 
}

@misc{agrawal2024pixtral12b,
      title={Pixtral 12B}, 
      author={Pravesh Agrawal and Szymon Antoniak and Emma Bou Hanna and Baptiste Bout and Devendra Chaplot and Jessica Chudnovsky and Diogo Costa and Baudouin De Monicault and Saurabh Garg and Theophile Gervet and Soham Ghosh and Amélie Héliou and Paul Jacob and Albert Q. Jiang and Kartik Khandelwal and Timothée Lacroix and Guillaume Lample and Diego Las Casas and Thibaut Lavril and Teven Le Scao and Andy Lo and William Marshall and Louis Martin and Arthur Mensch and Pavankumar Muddireddy and Valera Nemychnikova and Marie Pellat and Patrick Von Platen and Nikhil Raghuraman and Baptiste Rozière and Alexandre Sablayrolles and Lucile Saulnier and Romain Sauvestre and Wendy Shang and Roman Soletskyi and Lawrence Stewart and Pierre Stock and Joachim Studnia and Sandeep Subramanian and Sagar Vaze and Thomas Wang and Sophia Yang},
      year={2024},
      eprint={2410.07073},
      archivePrefix={arXiv},
      primaryClass={cs.CV},
      url={https://arxiv.org/abs/2410.07073}, 
}

@misc{openai2025gptoss120bgptoss20bmodel,
      title={gpt-oss-120b \& gpt-oss-20b Model Card}, 
      author={OpenAI},
      year={2025},
      eprint={2508.10925},
      archivePrefix={arXiv},
      primaryClass={cs.CL},
      url={https://arxiv.org/abs/2508.10925}, 
}

@misc{deepseekai2025deepseekr1incentivizingreasoningcapability,
      title={DeepSeek-R1: Incentivizing Reasoning Capability in LLMs via Reinforcement Learning}, 
      author={DeepSeek-AI},
      year={2025},
      eprint={2501.12948},
      archivePrefix={arXiv},
      primaryClass={cs.CL},
      url={https://arxiv.org/abs/2501.12948}, 
}

@inproceedings{liu2023llava,
    author      = {Liu, Haotian and Li, Chunyuan and Wu, Qingyang and Lee, Yong Jae},
    title       = {Visual Instruction Tuning},
    booktitle   = {NeurIPS},
    year        = {2023}
  }

@article{yue2023mmmu,
  url = {https://arxiv.org/abs/2311.16502},
  title={MMMU: A Massive Multi-discipline Multimodal Understanding and Reasoning Benchmark for Expert AGI},
  author={Yue, Xiang and Ni, Yuansheng and Zhang, Kai and Zheng, Tianyu and Liu, Ruoqi and Zhang, Ge and Stevens, Samuel and Jiang, Dongfu and Ren, Weiming and Sun, Yuxuan and others},
  journal={arXiv preprint arXiv:2311.16502},
  year={2023}
}

@article{yue2024mmmu,
  title={MMMU-Pro: A More Robust Multi-discipline Multimodal Understanding Benchmark},
  author={Xiang Yue and Tianyu Zheng and Yuansheng Ni and Yubo Wang and Kai Zhang and Shengbang Tong and Yuxuan Sun and Botao Yu and Ge Zhang and Huan Sun and Yu Su and Wenhu Chen and Graham Neubig},
  journal={arXiv preprint arXiv:2409.02813},
  year={2024}
}

@article{lu2023mathvista,
  url = {https://arxiv.org/abs/2310.02255},
  title={MathVista: Evaluating Mathematical Reasoning of Foundation Models in Visual Contexts},
  author={Lu, Pan and Bansal, Hritik and Xia, Tony and Liu, Jiacheng and Li, Chunyuan and Hajishirzi, Hannaneh and Cheng, Hao and Chang, Kai-Wei and Galley, Michel and Gao, Jianfeng},
  journal={arXiv preprint arXiv:2310.02255},
  year={2023}
}

@article{wang2024charxiv,
  url = {https://arxiv.org/abs/2406.18521},
  title={CharXiv: Charting Gaps in Realistic Chart Understanding in Multimodal Large Language Models},
  author={Wang, Zirui and Xia, Mengzhou and He, Luxi and Chen, Howard and Liu, Yitao and Zhu, Richard and Liang, Kaiqu and Wu, Xindi and Liu, Haotian and Malladi, Sadhika and others},
  journal={arXiv preprint arXiv:2406.18521},
  year={2024}
}

@misc{duan2024vlmevalkit,
  title={VLMEvalKit: An open-source toolkit for evaluating large
                  multi-modality models},
  author={Duan, Haodong and Fang, Xinyu and Yang, Junming and Zhao,
                  Xiangyu and Qiao, Yuxuan and Li, Mo and Agarwal, Amit and
                  Chen, Zhe and Chen, Lin and Liu, Yuan and Ma, Yubo and Sun,
                  Hailong and Zhang, Yifan and Lu, Shiyin and Wong, Tack Hwa
                  and Wang, Weiyun and Zhou, Peiheng and Li, Xiaozhe and Fu,
                  Chaoyou and Cui, Junbo and Chen, Jixuan and Song, Enxin and
                  Mao, Song and Ding, Shengyuan and Liang, Tianhao and Zhang,
                  Zicheng and Dong, Xiaoyi and Zang, Yuhang and Zhang, Pan and
                  Wang, Jiaqi and Lin, Dahua and Chen, Kai},
  journal={arXiv preprint	arXiv:2407.11691},
  year={2024},
  eprint={2407.11691},
  archivePrefix={arXiv},
  primaryClass={cs.CL},
  url={https://arxiv.org/abs/2407.11691}, 
}

@misc{yue2024mmmupro,
  url = {https://arxiv.org/abs/2409.02813},
  author = {Yue,  Xiang and Zheng,  Tianyu and Ni,  Yuansheng and Wang,  Yubo and Zhang,  Kai and Tong,  Shengbang and Sun,  Yuxuan and Yu,  Botao and Zhang,  Ge and Sun,  Huan and Su,  Yu and Chen,  Wenhu and Neubig,  Graham},
  title = {MMMU-Pro: A More Robust Multi-discipline Multimodal Understanding Benchmark},
  publisher = {arXiv},
  year = {2024}
}

@misc{xiao2024logicvista,
  url = {https://arxiv.org/abs/2407.04973},
  author = {Xiao,  Yijia and Sun,  Edward and Liu,  Tianyu and Wang,  Wei},
  title = {LogicVista: Multimodal LLM Logical Reasoning Benchmark in Visual Contexts},
  publisher = {arXiv},
  year = {2024}
}

@misc{wang2024mathvision,
  url = {https://arxiv.org/abs/2402.14804},
  author = {Wang,  Ke and Pan,  Junting and Shi,  Weikang and Lu,  Zimu and Zhan,  Mingjie and Li,  Hongsheng},
  title = {Measuring Multimodal Mathematical Reasoning with MATH-Vision Dataset},
  publisher = {arXiv},
  year = {2024}
}

@misc{zhang2024mathverse,
  url = {https://arxiv.org/abs/2403.14624},
  author = {Zhang,  Renrui and Jiang,  Dongzhi and Zhang,  Yichi and Lin,  Haokun and Guo,  Ziyu and Qiu,  Pengshuo and Zhou,  Aojun and Lu,  Pan and Chang,  Kai-Wei and Gao,  Peng and Li,  Hongsheng},
  title = {MathVerse: Does Your Multi-modal LLM Truly See the Diagrams in Visual Math Problems?},
  publisher = {arXiv},
  year = {2024}
}

@misc{chen2024mmstar,
  url = {https://arxiv.org/abs/2403.20330},
  author = {Chen,  Lin and Li,  Jinsong and Dong,  Xiaoyi and Zhang,  Pan and Zang,  Yuhang and Chen,  Zehui and Duan,  Haodong and Wang,  Jiaqi and Qiao,  Yu and Lin,  Dahua and Zhao,  Feng},
  title = {Are We on the Right Way for Evaluating Large Vision-Language Models?},
  publisher = {arXiv},
  year = {2024}
}

@misc{kembhavi2016ai2d,
  url = {https://arxiv.org/abs/1603.07396},
  author = {Kembhavi,  Aniruddha and Salvato,  Mike and Kolve,  Eric and Seo,  Minjoon and Hajishirzi,  Hannaneh and Farhadi,  Ali},
  title = {A Diagram Is Worth A Dozen Images},
  publisher = {arXiv},
  year = {2016}
}

@misc{fu2024blink,
  url = {https://arxiv.org/abs/2404.12390},
  author = {Fu,  Xingyu and Hu,  Yushi and Li,  Bangzheng and Feng,  Yu and Wang,  Haoyu and Lin,  Xudong and Roth,  Dan and Smith,  Noah A. and Ma,  Wei-Chiu and Krishna,  Ranjay},
  title = {BLINK: Multimodal Large Language Models Can See but Not Perceive},
  publisher = {arXiv},
  year = {2024}
}

@misc{openai2025gpt5systemcard,
  author       = {{OpenAI}},
  title        = {{GPT-5 System Card}},
  year         = {2025},
  url          = {https://cdn.openai.com/gpt-5-system-card.pdf},
  note         = {White paper},
}

@misc{xai2025grok4fastsystemcard,
  author       = {{X.AI Corp}},
  title        = {{Grok-4 Fast System Card}},
  year         = {2025},
  url          = {https://data.x.ai/2025-09-19-grok-4-fast-model-card.pdf},
  note         = {White paper},
}

@misc{bai2025qwen25vl,
  url = {https://arxiv.org/abs/2502.13923},
  author = {Bai,  Shuai and Chen,  Keqin and Liu,  Xuejing and Wang,  Jialin and Ge,  Wenbin and Song,  Sibo and Dang,  Kai and Wang,  Peng and Wang,  Shijie and Tang,  Jun and Zhong,  Humen and Zhu,  Yuanzhi and Yang,  Mingkun and Li,  Zhaohai and Wan,  Jianqiang and Wang,  Pengfei and Ding,  Wei and Fu,  Zheren and Xu,  Yiheng and Ye,  Jiabo and Zhang,  Xi and Xie,  Tianbao and Cheng,  Zesen and Zhang,  Hang and Yang,  Zhibo and Xu,  Haiyang and Lin,  Junyang},
  title = {Qwen2.5-VL Technical Report},
  publisher = {arXiv},
  year = {2025}
}

@inproceedings{tiwari-etal-2025-auto,
    title = "Auto-Cypher: Improving {LLM}s on Cypher generation via {LLM}-supervised generation-verification framework",
    author = "Tiwari, Aman  and
      Malay, Shiva Krishna Reddy  and
      Yadav, Vikas  and
      Hashemi, Masoud  and
      Madhusudhan, Sathwik Tejaswi",
    editor = "Chiruzzo, Luis  and
      Ritter, Alan  and
      Wang, Lu",
    booktitle = "Proceedings of the 2025 Conference of the Nations of the Americas Chapter of the Association for Computational Linguistics: Human Language Technologies (Volume 2: Short Papers)",
    month = apr,
    year = "2025",
    address = "Albuquerque, New Mexico",
    publisher = "Association for Computational Linguistics",
    url = "https://aclanthology.org/2025.naacl-short.53/",
    doi = "10.18653/v1/2025.naacl-short.53",
    pages = "623--640",
    ISBN = "979-8-89176-190-2"
}

@misc{xu2025wizardlmempoweringlargepretrained,
      title={WizardLM: Empowering large pre-trained language models to follow complex instructions}, 
      author={Can Xu and Qingfeng Sun and Kai Zheng and Xiubo Geng and Pu Zhao and Jiazhan Feng and Chongyang Tao and Qingwei Lin and Daxin Jiang},
      year={2025},
      eprint={2304.12244},
      archivePrefix={arXiv},
      primaryClass={cs.CL},
      url={https://arxiv.org/abs/2304.12244}, 
}

@article{yang2024qwen2,
  title={Qwen2 Technical Report},
  author={Yang, An and others},
  journal={arXiv preprint arXiv:2407.10671},
  year={2024},
  url={https://arxiv.org/abs/2407.10671}
}

@article{yang2024qwen2_5,
  title={Qwen2.5 Technical Report},
  author={Yang, An and others},
  journal={arXiv preprint arXiv:2412.15115},
  year={2024},
  url={https://arxiv.org/abs/2412.15115}
}

@article{deepseekr1,
  title={DeepSeek-R1: Incentivizing Reasoning Capability in LLMs via Reinforcement Learning},
  author={Ren, Zehui and others},
  journal={arXiv preprint arXiv:2501.12948},
  year={2025},
  url={https://arxiv.org/abs/2501.12948}
}

@article{gemini15,
  title={Gemini 1.5: Unlocking multimodal understanding across millions of tokens of context},
  author={{Gemini Team, Google}},
  journal={arXiv preprint arXiv:2403.05530},
  year={2024},
  url={https://arxiv.org/abs/2403.05530}
}

@techreport{gemini25,
  title={Gemini 2.5: Pushing the Frontier with Advanced Reasoning, Multimodality, Long Context},
  author={{Gemini Team, Google DeepMind}},
  institution={Google DeepMind},
  year={2025},
  url={https://storage.googleapis.com/deepmind-media/gemini/gemini_v2_5_report.pdf}
}

@misc{claude35sonnet,
  title={Introducing Claude 3.5 Sonnet},
  author={Anthropic},
  howpublished={\url{https://www.anthropic.com/news/claude-3-5-sonnet}},
  year={2024},
  note={Accessed 2025-09-29}
}

@misc{slam-distillation-from-r1,  
    author = {Sathwik Tejaswi Madhusudhan and Shruthan Radhakrishna and Jash Mehta and Toby Liang},  
    title = {Millions scale dataset distilled from R1-32b},  
    howpublished = {https://huggingface.co/datasets/ServiceNow-AI/R1-Distill-SFT},
    publisher = {SLAM - ServiceNow Language Models Lab} , 
    year = {2025}
}

@misc{kimiteam2025kimivltechnicalreport,
      title={{Kimi-VL} Technical Report}, 
      author={Kimi Team and Angang Du and Bohong Yin and Bowei Xing and Bowen Qu and Bowen Wang and Cheng Chen and Chenlin Zhang and Chenzhuang Du and Chu Wei and Congcong Wang and Dehao Zhang and Dikang Du and Dongliang Wang and Enming Yuan and Enzhe Lu and Fang Li and Flood Sung and Guangda Wei and Guokun Lai and Han Zhu and Hao Ding and Hao Hu and Hao Yang and Hao Zhang and Haoning Wu and Haotian Yao and Haoyu Lu and Heng Wang and Hongcheng Gao and Huabin Zheng and Jiaming Li and Jianlin Su and Jianzhou Wang and Jiaqi Deng and Jiezhong Qiu and Jin Xie and Jinhong Wang and Jingyuan Liu and Junjie Yan and Kun Ouyang and Liang Chen and Lin Sui and Longhui Yu and Mengfan Dong and Mengnan Dong and Nuo Xu and Pengyu Cheng and Qizheng Gu and Runjie Zhou and Shaowei Liu and Sihan Cao and Tao Yu and Tianhui Song and Tongtong Bai and Wei Song and Weiran He and Weixiao Huang and Weixin Xu and Xiaokun Yuan and Xingcheng Yao and Xingzhe Wu and Xinxing Zu and Xinyu Zhou and Xinyuan Wang and Y. Charles and Yan Zhong and Yang Li and Yangyang Hu and Yanru Chen and Yejie Wang and Yibo Liu and Yibo Miao and Yidao Qin and Yimin Chen and Yiping Bao and Yiqin Wang and Yongsheng Kang and Yuanxin Liu and Yulun Du and Yuxin Wu and Yuzhi Wang and Yuzi Yan and Zaida Zhou and Zhaowei Li and Zhejun Jiang and Zheng Zhang and Zhilin Yang and Zhiqi Huang and Zihao Huang and Zijia Zhao and Ziwei Chen},
      year={2025},
      eprint={2504.07491},
      archivePrefix={arXiv},
      primaryClass={cs.CV},
      url={https://arxiv.org/abs/2504.07491}, 
}

@misc{llama4,
	month = {4},
    author={Meta AI},
	title = {The Llama 4 herd: The beginning of a new era of natively multimodal AI innovation},
	year = {2025},
	url = {https://ai.meta.com/blog/llama-4-multimodal-intelligence/},
}

@misc{grattafiori2024llama3herdmodels,
      title={The Llama 3 Herd of Models}, 
      author={Aaron Grattafiori and Abhimanyu Dubey and Abhinav Jauhri and Abhinav Pandey and Abhishek Kadian and Ahmad Al-Dahle and Aiesha Letman and Akhil Mathur and Alan Schelten and Alex Vaughan and Amy Yang and Angela Fan and Anirudh Goyal and Anthony Hartshorn and Aobo Yang and Archi Mitra and Archie Sravankumar and Artem Korenev and Arthur Hinsvark and Arun Rao and Aston Zhang and Aurelien Rodriguez and Austen Gregerson and Ava Spataru and Baptiste Roziere and Bethany Biron and Binh Tang and Bobbie Chern and Charlotte Caucheteux and Chaya Nayak and Chloe Bi and Chris Marra and Chris McConnell and Christian Keller and Christophe Touret and Chunyang Wu and Corinne Wong and Cristian Canton Ferrer and Cyrus Nikolaidis and Damien Allonsius and Daniel Song and Danielle Pintz and Danny Livshits and Danny Wyatt and David Esiobu and Dhruv Choudhary and Dhruv Mahajan and Diego Garcia-Olano and Diego Perino and Dieuwke Hupkes and Egor Lakomkin and Ehab AlBadawy and Elina Lobanova and Emily Dinan and Eric Michael Smith and Filip Radenovic and Francisco Guzmán and Frank Zhang and Gabriel Synnaeve and Gabrielle Lee and Georgia Lewis Anderson and Govind Thattai and Graeme Nail and Gregoire Mialon and Guan Pang and Guillem Cucurell and Hailey Nguyen and Hannah Korevaar and Hu Xu and Hugo Touvron and Iliyan Zarov and Imanol Arrieta Ibarra and Isabel Kloumann and Ishan Misra and Ivan Evtimov and Jack Zhang and Jade Copet and Jaewon Lee and Jan Geffert and Jana Vranes and Jason Park and Jay Mahadeokar and Jeet Shah and Jelmer van der Linde and Jennifer Billock and Jenny Hong and Jenya Lee and Jeremy Fu and Jianfeng Chi and Jianyu Huang and Jiawen Liu and Jie Wang and Jiecao Yu and Joanna Bitton and Joe Spisak and Jongsoo Park and Joseph Rocca and Joshua Johnstun and Joshua Saxe and Junteng Jia and Kalyan Vasuden Alwala and Karthik Prasad and Kartikeya Upasani and Kate Plawiak and Ke Li and Kenneth Heafield and Kevin Stone and Khalid El-Arini and Krithika Iyer and Kshitiz Malik and Kuenley Chiu and Kunal Bhalla and Kushal Lakhotia and Lauren Rantala-Yeary and Laurens van der Maaten and Lawrence Chen and Liang Tan and Liz Jenkins and Louis Martin and Lovish Madaan and Lubo Malo and Lukas Blecher and Lukas Landzaat and Luke de Oliveira and Madeline Muzzi and Mahesh Pasupuleti and Mannat Singh and Manohar Paluri and Marcin Kardas and Maria Tsimpoukelli and Mathew Oldham and Mathieu Rita and Maya Pavlova and Melanie Kambadur and Mike Lewis and Min Si and Mitesh Kumar Singh and Mona Hassan and Naman Goyal and Narjes Torabi and Nikolay Bashlykov and Nikolay Bogoychev and Niladri Chatterji and Ning Zhang and Olivier Duchenne and Onur Çelebi and Patrick Alrassy and Pengchuan Zhang and Pengwei Li and Petar Vasic and Peter Weng and Prajjwal Bhargava and Pratik Dubal and Praveen Krishnan and Punit Singh Koura and Puxin Xu and Qing He and Qingxiao Dong and Ragavan Srinivasan and Raj Ganapathy and Ramon Calderer and Ricardo Silveira Cabral and Robert Stojnic and Roberta Raileanu and Rohan Maheswari and Rohit Girdhar and Rohit Patel and Romain Sauvestre and Ronnie Polidoro and Roshan Sumbaly and Ross Taylor and Ruan Silva and Rui Hou and Rui Wang and Saghar Hosseini and Sahana Chennabasappa and Sanjay Singh and Sean Bell and Seohyun Sonia Kim and Sergey Edunov and Shaoliang Nie and Sharan Narang and Sharath Raparthy and Sheng Shen and Shengye Wan and Shruti Bhosale and Shun Zhang and Simon Vandenhende and Soumya Batra and Spencer Whitman and Sten Sootla and Stephane Collot and Suchin Gururangan and Sydney Borodinsky and Tamar Herman and Tara Fowler and Tarek Sheasha and Thomas Georgiou and Thomas Scialom and Tobias Speckbacher and Todor Mihaylov and Tong Xiao and Ujjwal Karn and Vedanuj Goswami and Vibhor Gupta and Vignesh Ramanathan and Viktor Kerkez and Vincent Gonguet and Virginie Do and Vish Vogeti and Vítor Albiero and Vladan Petrovic and Weiwei Chu and Wenhan Xiong and Wenyin Fu and Whitney Meers and Xavier Martinet and Xiaodong Wang and Xiaofang Wang and Xiaoqing Ellen Tan and Xide Xia and Xinfeng Xie and Xuchao Jia and Xuewei Wang and Yaelle Goldschlag and Yashesh Gaur and Yasmine Babaei and Yi Wen and Yiwen Song and Yuchen Zhang and Yue Li and Yuning Mao and Zacharie Delpierre Coudert and Zheng Yan and Zhengxing Chen and Zoe Papakipos and Aaditya Singh and Aayushi Srivastava and Abha Jain and Adam Kelsey and Adam Shajnfeld and Adithya Gangidi and Adolfo Victoria and Ahuva Goldstand and Ajay Menon and Ajay Sharma and Alex Boesenberg and Alexei Baevski and Allie Feinstein and Amanda Kallet and Amit Sangani and Amos Teo and Anam Yunus and Andrei Lupu and Andres Alvarado and Andrew Caples and Andrew Gu and Andrew Ho and Andrew Poulton and Andrew Ryan and Ankit Ramchandani and Annie Dong and Annie Franco and Anuj Goyal and Aparajita Saraf and Arkabandhu Chowdhury and Ashley Gabriel and Ashwin Bharambe and Assaf Eisenman and Azadeh Yazdan and Beau James and Ben Maurer and Benjamin Leonhardi and Bernie Huang and Beth Loyd and Beto De Paola and Bhargavi Paranjape and Bing Liu and Bo Wu and Boyu Ni and Braden Hancock and Bram Wasti and Brandon Spence and Brani Stojkovic and Brian Gamido and Britt Montalvo and Carl Parker and Carly Burton and Catalina Mejia and Ce Liu and Changhan Wang and Changkyu Kim and Chao Zhou and Chester Hu and Ching-Hsiang Chu and Chris Cai and Chris Tindal and Christoph Feichtenhofer and Cynthia Gao and Damon Civin and Dana Beaty and Daniel Kreymer and Daniel Li and David Adkins and David Xu and Davide Testuggine and Delia David and Devi Parikh and Diana Liskovich and Didem Foss and Dingkang Wang and Duc Le and Dustin Holland and Edward Dowling and Eissa Jamil and Elaine Montgomery and Eleonora Presani and Emily Hahn and Emily Wood and Eric-Tuan Le and Erik Brinkman and Esteban Arcaute and Evan Dunbar and Evan Smothers and Fei Sun and Felix Kreuk and Feng Tian and Filippos Kokkinos and Firat Ozgenel and Francesco Caggioni and Frank Kanayet and Frank Seide and Gabriela Medina Florez and Gabriella Schwarz and Gada Badeer and Georgia Swee and Gil Halpern and Grant Herman and Grigory Sizov and Guangyi and Zhang and Guna Lakshminarayanan and Hakan Inan and Hamid Shojanazeri and Han Zou and Hannah Wang and Hanwen Zha and Haroun Habeeb and Harrison Rudolph and Helen Suk and Henry Aspegren and Hunter Goldman and Hongyuan Zhan and Ibrahim Damlaj and Igor Molybog and Igor Tufanov and Ilias Leontiadis and Irina-Elena Veliche and Itai Gat and Jake Weissman and James Geboski and James Kohli and Janice Lam and Japhet Asher and Jean-Baptiste Gaya and Jeff Marcus and Jeff Tang and Jennifer Chan and Jenny Zhen and Jeremy Reizenstein and Jeremy Teboul and Jessica Zhong and Jian Jin and Jingyi Yang and Joe Cummings and Jon Carvill and Jon Shepard and Jonathan McPhie and Jonathan Torres and Josh Ginsburg and Junjie Wang and Kai Wu and Kam Hou U and Karan Saxena and Kartikay Khandelwal and Katayoun Zand and Kathy Matosich and Kaushik Veeraraghavan and Kelly Michelena and Keqian Li and Kiran Jagadeesh and Kun Huang and Kunal Chawla and Kyle Huang and Lailin Chen and Lakshya Garg and Lavender A and Leandro Silva and Lee Bell and Lei Zhang and Liangpeng Guo and Licheng Yu and Liron Moshkovich and Luca Wehrstedt and Madian Khabsa and Manav Avalani and Manish Bhatt and Martynas Mankus and Matan Hasson and Matthew Lennie and Matthias Reso and Maxim Groshev and Maxim Naumov and Maya Lathi and Meghan Keneally and Miao Liu and Michael L. Seltzer and Michal Valko and Michelle Restrepo and Mihir Patel and Mik Vyatskov and Mikayel Samvelyan and Mike Clark and Mike Macey and Mike Wang and Miquel Jubert Hermoso and Mo Metanat and Mohammad Rastegari and Munish Bansal and Nandhini Santhanam and Natascha Parks and Natasha White and Navyata Bawa and Nayan Singhal and Nick Egebo and Nicolas Usunier and Nikhil Mehta and Nikolay Pavlovich Laptev and Ning Dong and Norman Cheng and Oleg Chernoguz and Olivia Hart and Omkar Salpekar and Ozlem Kalinli and Parkin Kent and Parth Parekh and Paul Saab and Pavan Balaji and Pedro Rittner and Philip Bontrager and Pierre Roux and Piotr Dollar and Polina Zvyagina and Prashant Ratanchandani and Pritish Yuvraj and Qian Liang and Rachad Alao and Rachel Rodriguez and Rafi Ayub and Raghotham Murthy and Raghu Nayani and Rahul Mitra and Rangaprabhu Parthasarathy and Raymond Li and Rebekkah Hogan and Robin Battey and Rocky Wang and Russ Howes and Ruty Rinott and Sachin Mehta and Sachin Siby and Sai Jayesh Bondu and Samyak Datta and Sara Chugh and Sara Hunt and Sargun Dhillon and Sasha Sidorov and Satadru Pan and Saurabh Mahajan and Saurabh Verma and Seiji Yamamoto and Sharadh Ramaswamy and Shaun Lindsay and Shaun Lindsay and Sheng Feng and Shenghao Lin and Shengxin Cindy Zha and Shishir Patil and Shiva Shankar and Shuqiang Zhang and Shuqiang Zhang and Sinong Wang and Sneha Agarwal and Soji Sajuyigbe and Soumith Chintala and Stephanie Max and Stephen Chen and Steve Kehoe and Steve Satterfield and Sudarshan Govindaprasad and Sumit Gupta and Summer Deng and Sungmin Cho and Sunny Virk and Suraj Subramanian and Sy Choudhury and Sydney Goldman and Tal Remez and Tamar Glaser and Tamara Best and Thilo Koehler and Thomas Robinson and Tianhe Li and Tianjun Zhang and Tim Matthews and Timothy Chou and Tzook Shaked and Varun Vontimitta and Victoria Ajayi and Victoria Montanez and Vijai Mohan and Vinay Satish Kumar and Vishal Mangla and Vlad Ionescu and Vlad Poenaru and Vlad Tiberiu Mihailescu and Vladimir Ivanov and Wei Li and Wenchen Wang and Wenwen Jiang and Wes Bouaziz and Will Constable and Xiaocheng Tang and Xiaojian Wu and Xiaolan Wang and Xilun Wu and Xinbo Gao and Yaniv Kleinman and Yanjun Chen and Ye Hu and Ye Jia and Ye Qi and Yenda Li and Yilin Zhang and Ying Zhang and Yossi Adi and Youngjin Nam and Yu and Wang and Yu Zhao and Yuchen Hao and Yundi Qian and Yunlu Li and Yuzi He and Zach Rait and Zachary DeVito and Zef Rosnbrick and Zhaoduo Wen and Zhenyu Yang and Zhiwei Zhao and Zhiyu Ma},
      year={2024},
      eprint={2407.21783},
      archivePrefix={arXiv},
      primaryClass={cs.AI},
      url={https://arxiv.org/abs/2407.21783}, 
}
\newpage
\section{Contributions and Acknowledgments}
\label{sec:contributors}

\textbf{Core Contributors}\\[0.3em]
Shruthan Radhakrishna, Aman Tiwari,
Aanjaneya Shukla, Masoud Hashemi, Rishabh Maheshwary, Shiva Krishna Reddy Malay, Jash Mehta, Pulkit Pattnaik, Saloni Mittal,
Khalil Slimi, Kelechi Ogueji, Akintunde Oladipo, Soham Parikh, Oluwanifemi Bamgbose  \\ [0.35em]

\textbf{Secondary Contributors}\\[0.3em]
Toby Liang, Ahmed Masry, Khyati Mahajan, Sai Rajeswar Mudumba, Vikas Yadav\\[0.35em]

\renewcommand{\arraystretch}{1.3} 
\textbf{Leads \& Management}\\
\begin{table}[h!]
\centering
\begin{tabular}{|p{0.4\textwidth}|p{0.6\textwidth}|}
\hline
Sathwik Tejaswi Madhusudhan & Technical co-lead (Mid-Training and Post-Training) \\ \hline
Torsten Scholak & Technical co-lead (Pre-Training and Architecture) \\ \hline
Sagar Davasam & Applied Research Manager \\ \hline
Srinivas Sunkara & VP, Applied Research \\ \hline
Nicholas Chapados & VP, AI Research \\ \hline
\end{tabular}
\end{table}

\textbf{Upstream Contributors}\\[0.3em]
We gratefully acknowledge the contributors whose work on Apriel-Nemotron-15B-Thinker and related projects was reused as part of the current work:
\\

Gopal Sarda, Anil Turkkan, Shashank Maiya, Dhruv Jhamb, Jishnu S Nair, Akshay Kalkunte, Bidyapati Pradhan, Surajit Dasgupta, Jaykumar Kasundra, Anjaneya Praharaj, Sourabh Surana, Lakshmi Sirisha Chodisetty, Abhigya Verma, Abhishek Bharadwaj, Nandhakumar Kandasamy, Naman Gupta, Segan Subramanian \\[0.25em]

We also thank Anil Madamala for leading evaluations and benchmarking, and Segan Subramanian and Vipul Mittal for leading data infrastructure.

\end{document}